\documentclass{article}

\usepackage{graphicx}
\usepackage{amsmath}
\usepackage{amssymb}
\usepackage{algorithm}
\usepackage{algorithmicx}
\usepackage{algpseudocode}
\usepackage{subfig}
\usepackage{tikz}

\usepackage{ogonek}

\usetikzlibrary{shapes}
\usetikzlibrary{arrows}
\tikzset{
    >=stealth',
}
\newcommand{\cen}{Ce}%centromere
\newcommand{\coa}{Co}%coarse speckled
\newcommand{\cyt}{Cy}%cytoplasmatic
\newcommand{\fin}{Fi}%fine_speckled  
\renewcommand{\hom}{Ho}%homogeneous    
\newcommand{\nuc}{Nu}%nucleolar 

\newcommand{\centxt}{centromere}
\newcommand{\coatxt}{coarse speckled}
\newcommand{\cyttxt}{cytoplasmatic}
\newcommand{\fintxt}{fine speckled}  
\newcommand{\homtxt}{homogeneous}  
\newcommand{\nuctxt}{nucleolar}

\newcommand{\gauss}[4]{%
\draw[smooth,domain=0:5,samples=50] plot (\x,{#2*exp((0-(\x-#1)^2)*#4)});
\node[anchor=south west] at (#1,#2) {#3}; }

\begin{document}

%% Title, authors and addresses

%% use the tnoteref command within \title for footnotes;
%% use the tnotetext command for the associated footnote;
%% use the fnref command within \author or \address for footnotes;
%% use the fntext command for the associated footnote;
%% use the corref command within \author for corresponding author footnotes;
%% use the cortext command for the associated footnote;
%% use the ead command for the email address,
%% and the form \ead[url] for the home page:
%%
%% \title{Title\tnoteref{label1}}
%% \tnotetext[label1]{}
%% \author{Name\corref{cor1}\fnref{label2}}
%% \ead{email address}
%% \ead[url]{home page}
%% \fntext[label2]{}
%% \cortext[cor1]{}
%% \address{Address\fnref{label3}}
%% \fntext[label3]{}

\title{Quantile Representation for Indirect Immunofluorescence Image Classification }

\author{David M. J. Tax, Veronika Cheplygina, Marco Loog}

\maketitle

\begin{abstract}
In the diagnosis of autoimmune diseases, an important task is to classify images of slides containing several HEp-2 cells. All cells from one slide share the same label, and by classifying cells from one slide independently, some information on the global image quality and intensity is lost. Considering one whole slide as a collection (a bag) of feature vectors, however, poses the problem of how to handle this bag. A simple, and surprisingly effective, approach is to summarize the bag of feature vectors by a few quantile values per feature. This characterizes the full distribution of all instances, thereby assuming that all instances in a bag are informative. This representation is particularly useful when each bag contains many feature vectors, which is the case in the classification of the immunofluorescence images. Experiments on the classification of indirect immunofluorescence images show the usefulness of this approach. 
\end{abstract}

\section{Introduction}

Anti-nuclear antibodies (ANAs) are autoantibodies that bind to cell nuclei. In healthy individuals, the immune system produces antibodies against foreign proteins or antigens, but not against human proteins (autoantigens). However, in individuals suffering from autoimmune disorders, antibodies for autoantigens are produced as well. Therefore, ANAs are important for diagnostic purposes and different types of ANAs are indicative of different illnesses~\cite{kavanaugh2000guidelines}. 

A common test for detecting ANAs is indirect immunofluorescence (IIF) with slides of HEp-2 cells. ANAs bind to the cell nuclei, and the type of ANAs (and therefore type of illness) can be distinguished by examining the staining pattern of the cells. Problems such as the presence of noise, low level of standardization and errors in interpretation all lead to uncertainty in the diagnoses. Due to these challenges, computer aided diagnosis has been suggested as an alternative to examining the staining patterns~\cite{perner2002mining}. 
%http://mivia.unisa.it/hep2contest/background.shtml

Recently, an annotated dataset of six types of staining patterns from 28 subjects has been provided~\cite{foggia2010early} to facilitate the development of classifiers for the HEp-2 staining patterns. The dataset consists of 28 images of slides (one per subject), where each image contains several cells of the same class. A competition organized at the International Conference of Pattern Recognition 2012 revealed that there is large variability between cells of the same class, but across different subjects, leading to poor generalization. Our goal is to address this variability using a multiple instance learning~\cite{maron1998framework} approach.

In multiple instance learning an object is represented by a collection of feature vectors~\cite{DieLatLaz1997}. Typically, this set is called a bag, and the feature vectors inside a bag are called the instances. This is an extension to the standard pattern recognition approach where objects are represented by a single feature vector only. By using a set of feature vectors, the representational capacity is enriched, which potentially increases the discriminability between classes. The obvious drawback is that for the classification of an object, the full set of feature vectors has to be taken into account. Depending on assumptions on how to deal with the set of feature vectors, several multiple instance learning approaches have been proposed. One branch of classifiers focuses on finding the single most informative feature vector (approaches such as miSVM~\cite{AndTsoHof2003}, Diverse Density~\cite{maron1998framework}, and Expectation-Maximization~\cite{ZhaGol2002} fall into this category). Another branch focuses on describing the full distribution of the set of feature vectors (with approaches such as Citation-$k$NN~\cite{WanZuc2000}, MILES~\cite{CheBiWan2007}, kernels~\cite{gartner2002multi} and bag dissimilarities~\cite{tax2011bag,cheplygina2012does}).

When classifiying HEp-2 cells, the slides are the bags and the individual cells are the instances. Because cells in the same slide have the same label, selecting a single most informative cell per slide does not seem appropriate. Therefore we focus on approaches that describe the full distribution of the bags. We propose representing a bag by the quantiles of its instance distribution, an approach that is a generalization of the minimum-maximum representation in \cite{gartner2002multi}. In the minimum-maximum representation each bag is represented by the minimum and maximum feature value that appears in the bag. When the number of instances in a bag is large, these minima and maxima may be heavily influenced by outliers. By selecting appropriate quantiles, noisy feature values are avoided.

In Section \ref{sec:mil} a short explanation of learning on sets of feature vectors is given. Next, in Section \ref{sec:dataset} the dataset and its preprocessing is explained, followed by the proposed approach for the classification of HEp-2 cells in Section \ref{sec:proposal}. Finally, in Sections \ref{sec:exp} and \ref{sec:concl} experiments are shown and conclusions are drawn.

\section{Learning with sets}\label{sec:mil}

In multiple instance learning (MIL)~\cite{maron1998framework} and group-based classification (GBC)~\cite{samsudin2010nearest}, objects are represented as sets of feature vectors, rather than single feature vectors only. In MIL, labels are available only at a coarse level, for a set (\emph{bag}) of feature vectors (\emph{instances}). Both training and test objects are bags. In GBC, only the test objects are bags, with the added information that all instances inside one bag are from the same class.

In the case of HEp-2 cell classification, we can consider slides as bags, and individual cells as the instances. There are two advantages to doing so. 

In the training phase, a clear advantage of MIL is that learning is possible even if only coarse bag labels are availble. Although in the cell classification problem, the instance labels are available as well, there is an another advantage to using MIL in the training phase. In the case that the cells within one image are not independent, considering these cells jointly can be more informative than considering all cells individually. 

\begin{figure}[ht]
\centering
    
    \begin{tikzpicture}
    \tikzstyle{cl1}=[circle,draw,scale=0.6, minimum width=1pt, fill=black]
    
    \draw[->] (0,0) -- (5,0) node[right] {$x$};
    %\draw[->] (0,0) -- (0,2) node[above] {$p(x)$};
	\gauss{1.5}{1.5}{$p(x|\omega_1$)}{1}
	\gauss{3}{1.5}{$p(x|\omega_2$)}{1}
    
    \node[cl1] (x1) at (3,0) {}; 
    \node[cl1] (x1) at (1.8,0) {}; 
    \node[cl1] (x1) at (4.2,0) {}; 
    
	\end{tikzpicture}	
		
\caption{1-dimensional binary classification problem. The shaded circles are from the $\omega_2$ class. The added information that the circles are all of the same class helps to reduce the classification error.}
\label{fig:milexample}
\end{figure}
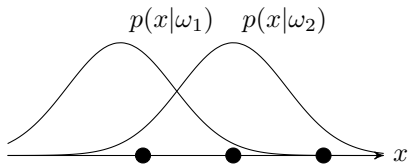

In the testing phase, considering several cells jointly can also help to improve performance. This is illustrated in the following example. Consider the 1-dimensional binary classification problem in Fig. \ref{fig:milexample}, and assume that we have found the Bayes optimal classifier. The shaded circles are the test set, and their true labels are $\omega_2$. If we were to classify these instances independently, the error would be equal to $\frac{1}{3}$, because the leftmost object will be misclassified to class $\omega_1$. However, with the added information that these instances belong to a bag of objects from the same class, we could apply a majority voting combination to classify the bag as $\omega_2$, and propagate the label to all the individual instances, reducing the error to 0. 

%Old:
%In the case of HEp-2 cell classification, this would mean that labels are available only for complete slides, but the individual cells are not labeled. In GBC, only the test objects are bags, with the added information that all instances inside this bag are from the same class. 

%A clear advantage of MIL is that learning is possible even with such coarse bag labels. Initially, it might seem that nothing can be gained in GBC where the fine level, instance labels are also present. However, this is not always the case, as illustrated in the following example. 

Considering bags of instances rather than single instances leads to a different representation than in regular supervised learning methods. One object is represented by a bag $B_i  = \{\mathbf{x}_{ik}| k=1,...,n_i\}, \,\mathbf{x} \in \mathbb{R}^d$ where $n_i$ is the number of instances in the bag, and which may be different for different bags. However, the disadvantage here is that supervised learning methods cannot be applied directly. 

One possibility to both consider information from all instances, and to retain a single feature vector representation, is to represent a bag as a distribution in instance space~\cite{gartner2002multi,tax2012distribution}. In \cite{gartner2002multi}, the so-called Minimax kernel represents each bag by the minimum and maximum values of its features, resulting in a $2d$-dimensional feature vector:

\begin{equation}
\mathbf{b}_i = [\min_k((\mathbf{x}_{ik})_1), \max_k((\mathbf{x}_{ik})_1), \ldots,  \min_k((\mathbf{x}_{ik})_d), \max_k((\mathbf{x}_{ik})_d)]
\end{equation}
where $(\mathbf{x})_l$ selects the $l$-th element of vector $\mathbf{x}$.

The same principle can be applied to other quantiles of the instances. 
When we again consider a bag $B_i$ with instances $\{\mathbf{x}_{ik}| k=1,...,n_i\}$, we obtain the $q$-th quantile for feature $l$ by first sorting all values of one feature, and then selecting the appropriate value from the sorted list:
\begin{equation}
b_{l,q}(\mathbf{x}_i) = (\mathbf{s}_l)_{\lfloor q\cdot n_i \rfloor},\quad \mathbf{s}_l = \textrm{sort}(\{(\mathbf{x}_{ik})_l\})
\end{equation}

In principle this quantile representation can be computed on bags of any size. When the number of instances per bag is small, many of these quantiles will coincide. Therefore the number of quantiles is typically chosen to be reasonably small, as a fraction of the average number of instances in a bag. For problems where the number of instances per bag is large, the extreme quantiles ($q=0$ or $q=1$, or the minimum or maximum feature value) can become noisy and in these situations these extreme quantiles should be avoided. For smaller bag sizes the minimum and maximum are often used and show good performance.

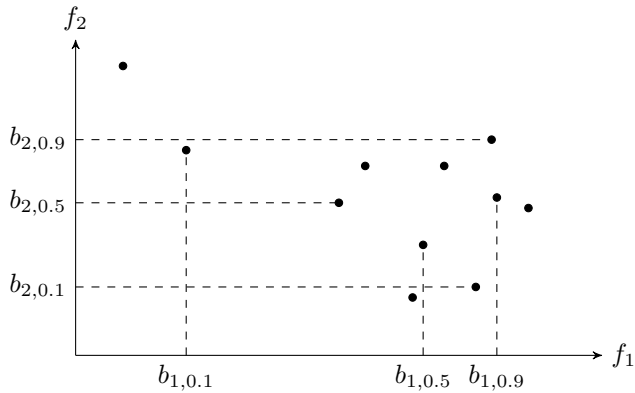
\begin{figure}[ht]
\centering\begin{tikzpicture}[scale = 0.7]
\tikzstyle{cl1}=[circle,draw,scale=0.3, minimum width=1pt, fill=black]
\draw[->] (0,0) -- (10,0) node[right] {$f_1$};
\draw[->] (0,0) -- (0,6) node[above] {$f_2$};
\node[cl1] at (2.1,3.9) {};
\node[cl1] at (0.9,5.5) {};
\node[cl1] at (8.0,3.0) {};
\node[cl1] at (7.0,3.6) {};
\node[cl1] at (8.6,2.8) {};
\node[cl1] at (7.6,1.3) {};
\node[cl1] at (7.9,4.1) {};
\node[cl1] at (5.0,2.9) {};
\node[cl1] at (5.5,3.6) {};
\node[cl1] at (6.4,1.1) {};
\node[cl1] at (6.6,2.1) {};
%\draw[dashed] (0.9,0.0) node[below] {$b_{1,0}$} -- (0.9,5.5); %min
\draw[dashed] (2.1,0.0) node[below] {$b_{1,0.1}$} -- (2.1,3.9);  %10%
\draw[dashed] (6.6,0.0) node[below] {$b_{1,0.5}$} -- (6.6,2.1);  %median
\draw[dashed] (8.0,0.0) node[below] {$b_{1,0.9}$} -- (8.0,3.0);  %90%
%\draw[dashed] (8.6,0.0) node[below] {$b_{1,1}$} -- (8.6,2.8); %max
%\draw[dashed] (0.0,1.1) node[left] {$b_{2,0}$} -- (6.4,1.1);   %min
\draw[dashed] (0.0,1.3) node[left] {$b_{2,0.1}$} -- (7.6,1.3);   %10%
\draw[dashed] (0.0,2.9) node[left] {$b_{2,0.5}$} -- (5.0,2.9);   %median
\draw[dashed] (0.0,4.1) node[left] {$b_{2,0.9}$} -- (7.9,4.1);   %90%
%\draw[dashed] (0.0,5.5) node[left] {$b_{2,1}$} -- (0.9,5.5);   %max
\end{tikzpicture}
\caption{Scatterplot of one bag with 11 instances. From this bag the 10\%-, 50\% and 90\%-quantiles are determined for each feature. This results in a 6D feature vector $[b_{1,0.1},b_{1,0.5},b_{1,0.9},b_{2,0.1},b_{2,0.5},b_{2,0.9}]$ for this bag.}
\label{fig:quantexample}
\end{figure}

To illustrate the quantile representation, a scatterplot of a single bag with 11 instances is shown in Fig. \ref{fig:quantexample}. For each of the features the 10\%-quantile the 50\%-quantile (the median) and the 90\%-quantile are computed. For this 2D feature space and these three different quantiles this results in a 6-dimensional feature vector for each bag.

Note that the correlation between the features is lost in this representation. The instance that has the highest feature value for the first feature is typically not the same instance with the highest feature value for the second feature. On the other hand, by considering each marginal distribution independently it is avoided that correlations and higher order statistics have to be estimated. For larger feature dimensionalities and finite datasets this is very challenging and should be avoided to not suffer from the curse of dimensionality.

\section{Dataset}\label{sec:dataset}

There are two different datasets available: a training set containing 721 cells scanned from 14 different images and a test set containing 734 cells scanned from 14 different images. All cells from one image share the same label. In six images all cells are ``centromere'', in five images they are labeled ``coarse speckled'' and five ``homogeneous''. For the classes ``cytoplasmatic'', ``fine speckled'', and ``nucleolar'' there are four images each. The images contain 52 cells on average, with a minimum of 13, and a maximum of 119 cells per image.

For each cell the green channel intensity is normalized using histogram equalization. On these normalized images Gabor features are computed. A Gabor filter is the product of a Gaussian kernel and a sinusoid:
\begin{equation}
g(x,y)=\exp\left(-\frac{x'^2+y'^2}{2\sigma^2}\right)\cos\left(2\pi\frac{x'}{\lambda}\right)
\end{equation}
where $x'=x\cos\theta+y\sin\theta$ and $y'=-x\sin\theta+y\cos\theta$. The parameter $\sigma$ defines the scale of the filter, the parameter $\theta$ determines the direction in which the cosine intensities vary, and $\lambda$ determines the wavelength of the cosine. In the experiments several values for these parameters are used: $\sigma=1,2,3,5,7$, $\theta=0,\pi/4,\pi/2,3\pi/4$ and $\lambda=0.05,0.1,0.2,0.3$. In total 80 filtered images are obtained per cell. 

To obtain a single feature vector for each cell, some simple statistics are derived from the image: the mean of the absolute values, the maximum, and the variance over all pixels within the mask. Next to that, the maximum, the variance and the mean of the image intensity are added. Finally, to obtain a rotational invariant representation of the cells, the outputs of the four different directions $\theta$ are averaged. This results in a $3\times 80/4 + 3 = 63$ dimensional feature vector. 
The results in section \ref{sec:exp} show that this representation is expressive enough for a reasonable cell classification performance.

\section{Proposed approach}\label{sec:proposal}

We propose to formulate the problem in the multiple instance learning setting, therefore learning on images rather than individual cells.

Our procedure is shown in Algorithm \ref{alg1}. In our implementation, the procedure $extractFeatures$ extracts the Gabor features as described in Section \ref{sec:dataset}, $quantileRepresentation$ represents each bag by a distribution of its instances and $trainClassifier$ and $testClassifier$ uses a logistic classifier. However, in principle these implementations can be replaced by other features, distributions and classifiers.

\begin{algorithm}
\caption{Classification procedure}\label{alg1}
\begin{algorithmic}[1]
\Procedure{ClassifyCells}{$trainSlides, testSlides, trainIndex, testIndex$}%\Comment{Text}

\For{$image_i \in trainImages \cup testImages$ }
    \For{$cell_j \in image_i$ }
    \State $cellData_{i,j} \gets extractFeatures(cell_j)$
    \EndFor
\State $slideData_i \gets quantileRepresentation(cellData_{i,:})$\;
\EndFor

\State $classifier \gets trainClassifier(slideData, trainIndex)$\;
\State $testLabels \gets testClassifier(slideData,testIndex,classifier)$\;

\State \textbf{return} $testLabels$ %\Comment{Labels for slides}
\EndProcedure

\end{algorithmic}
\end{algorithm}

We examined several quantiles for the cell data: minimum, 10\%, 50\%, 90\% and maximum. The minimum quantile turned out too noisy, but the 10\%-quantile gave very good results. On the other hand, the maximum performed better than the 90\% quantile, therefore we used the 10\%-, 50\%- and 100\%-quantiles as the basis for our approach.

Adding other quantiles to this list generally deteriorated the performance, unless a quantile was selected that was close to one of the already chosen extremes, such as 11\%- or 99\%. Our final implementation includes the 10\%-, 11\%-, 50\%- and 100\%-quantiles.

%Why? 

\section{Experiments}\label{sec:exp}

First some results are shown to evaluate the standard feature-based approach using the Gabor features. Several classifiers have been trained on the individual cells, like the $k$-nearest neighbor, the standard support vector classifier or the logistic classifier. The best performing classifier (with a small margin) is the $L_1$ support vector classifier (or the Liknon \cite{BhaGraRizRadMolJorBisMi2003}). This classifier handles the multi-class problem by training six one-vs-all classifiers and combining the output using a max combiner. The performance on cell level and on image level is computed, where the label of an image is computed by majority voting over all predictions of the individual cells in that image.
 
We then present results of our quantile-based multiple instance learning approach, where whole bags are considered as train and test objects. Each bag is represented by the quantile representation. The logistic classifier is trained on the bags. The performance on bag level can therefore be derived directly, the performance on cell level is computed by propagating the image label to all individual cells in that image. 

The evaluation is the same for both approaches. The evaluation contains two performance measures: the first is the classification error on the pre-defined test set, the second is the averaged classification error using leave-one-image-out cross-validation. Because there are 28 images in total, this means a 28-fold cross-validation. Next to the classification errors, the confusion matrices are reported as well.

All experiments are performed in MATLAB\textsuperscript{\textregistered} with the PRTools toolbox~\cite{prtools}. 

\subsection{Cell level}

%Train a Liknon/L1 svm. %Hmm, for bags this is not very good... logistic better-2
\begin{table}[!ht]
\caption{The confusion matrices and classification performances using the $L_1$ support vector classifier on the eight HEp cell classes: \cen:``centromere'', \coa:``coarse speckled'', \cyt:``cytoplasmatic'', \fin:``fine speckled'', \hom:``homogeneous'', \nuc:``nucleolar''. Top row shows results on the predefined train and test set, the bottom row the results on 28-fold cross-validation.}
\label{tab:featbased}
\small
\begin{tabular}{cc}
\hline
\multicolumn{2}{c}{Split in predefined train and test set}\\
\hline
cell level evaluation & image level evaluation\\
\begin{tabular}{l|*{6}{c}}
& \multicolumn{6}{c}{estimated class} \\
true & \cen & \coa & \cyt & \fin & \hom & \nuc \\ 
 \hline 
      \cen &  94 &   0 &   0 &   1 &   4 &  50 \\
      \coa &   9 &  29 &   6 &  49 &   3 &   5 \\
      \cyt &   0 &   3 &  47 &   0 &   1 &   0 \\
      \fin &  25 &   8 &   0 &  39 &  36 &   6 \\
      \hom &   1 &   2 &   0 &  30 & 141 &   6 \\
      \nuc &  16 &   1 &   0 &   1 &  17 & 104 \\
\end{tabular}
&
\begin{tabular}{l|*{6}{c}}
& \multicolumn{6}{c}{estimated class} \\
true & \cen & \coa & \cyt & \fin & \hom & \nuc \\ 
 \hline 
      \cen &   2 &   0 &   0 &   0 &   0 &   1 \\
      \coa &   0 &   1 &   0 &   2 &   0 &   0 \\
      \cyt &   0 &   0 &   2 &   0 &   0 &   0 \\
      \fin &   0 &   0 &   0 &   1 &   1 &   0 \\
      \hom &   0 &   0 &   0 &   0 &   2 &   0 \\
      \nuc &   0 &   0 &   0 &   0 &   0 &   2 \\
\end{tabular}
\\
61.85\% correct & 71.43\% correct \\
\hline
\multicolumn{2}{c}{28-fold cross-validation}\\
\hline
cell level & image level \\
\begin{tabular}{l|*{6}{c}}
& \multicolumn{6}{c}{estimated class} \\
true & \cen & \coa & \cyt & \fin & \hom & \nuc \\ 
 \hline 
\cen & 239 &  36 &   0 &  37 &   4 &  41 \\
\coa &  54 &  77 &   5 &  58 &   6 &  10 \\
\cyt &   8 &   4 &  92 &   2 &   0 &   3 \\
\fin &  25 &  61 &   0 &  35 &  87 &   0 \\
\hom &   2 &   4 &   4 &  33 & 281 &   6 \\
\nuc &  39 &  16 &   1 &  14 &   8 & 163 \\
\end{tabular}
&
\begin{tabular}{l|*{6}{c}}
& \multicolumn{6}{c}{estimated class} \\
true & \cen & \coa & \cyt & \fin & \hom & \nuc \\ 
 \hline 
\cen &   5 &   0 &   0 &   1 &   0 &   0 \\
\coa &   1 &   1 &   0 &   3 &   0 &   0 \\
\cyt &   0 &   0 &   4 &   0 &   0 &   0 \\
\fin &   0 &   1 &   0 &   1 &   2 &   0 \\
\hom &   0 &   0 &   0 &   0 &   5 &   0 \\
\nuc &   0 &   0 &   0 &   0 &   0 &   4 \\
\end{tabular}
\\
60.96\% correct & 71.43\% correct \\
\end{tabular}
\end{table}

The top row of Table \ref{tab:featbased} shows the confusion matrices and the classification performance on the predefined train and test set.
In the top left, we evaluate the individual cells in the test set. The confusion matrix shows that there is large confusion between the classes ``coarse speckled'' and ``fine speckled'', between ``fine speckled'' and ``homogeneous'', and between ``fine speckled'' and ``centromere''. The errors are not symmetric though, from the ``coarse'' class many cells are assigned to the ``fine'' class, but from the ``fine'' class many are also assigned to the homogeneous class. Apparently, there is a gradual change from ``homegeneous'', via ``fine'' to ``coarse'', and the decision boundaries between the classes are not exactly in between the classes. Around 62\% of the cells in the test set is classified correctly.
In the top right, the confusion matrix and performance is given for the image level classification. The classification accuracy improves a bit from $61\%$ to $71\%$, but still there is a large confusion between ``coarse speckled'' and ``fine speckled''.

The second row of Table \ref{tab:featbased} shows the same setup with results for 28-fold cross-validation over all images. 
The overall performances do not differ a lot with the predefined train test split, but the confusion matrices show a similar confusion pattern between the classes.

\begin{table}[!ht]
\centering
\caption{Classification performance for each individual image. From left to right: the image number, the true class label, the number of cells that is correctly classified, and the fraction of cells that is correctly classified.}
\label{tab:imageperf}
\small
\begin{tabular}{rlrr}
 & true label & corr. & fraction\\
\hline
 1 & homogeneous     &  54 & 88.5\%\\
 2 & fine speckled   &   5 & 10.4\%\\
 3 & centromere      &  58 & 65.2\%\\
 4 & nucleolar       &  28 & 42.4\%\\
 5 & homogeneous     &  42 & 89.4\%\\
 6 & coarse speckled &  22 & 32.4\%\\
 7 & centromere      &  47 & 83.9\%\\
 8 & nucleolar       &  27 & 48.2\%\\
 9 & fine speckled   &   1 & 2.2\%\\
10 & coarse speckled &   9 & 27.3\%\\
11 & coarse speckled &  23 & 56.1\%\\
12 & coarse speckled &  20 & 40.8\%\\
13 & centromere      &  40 & 87.0\%\\
14 & centromere      &  27 & 42.9\%\\
\end{tabular}
\begin{tabular}{rlrr}
 & true label & corr. & fraction\\
\hline
15 & fine speckled   &  24 & 38.1\%\\
16 & centromere      &  35 & 92.1\%\\
17 & coarse speckled &   3 & 15.8\%\\
18 & homogeneous     &  31 & 73.8\%\\
19 & centromere      &  32 & 49.2\%\\
20 & nucleolar       &  43 & 93.5\%\\
21 & homogeneous     &  42 & 68.9\%\\
22 & homogeneous     & 112 & 94.1\%\\
23 & fine speckled   &   5 & 9.8\%\\
24 & nucleolar       &  65 & 89.0\%\\
25 & cytoplasmatic   &  13 & 54.2\%\\
26 & cytoplasmatic   &  31 & 91.2\%\\
27 & cytoplasmatic   &  36 & 94.7\%\\
28 & cytoplasmatic   &  12 & 92.3\%\\
\end{tabular}
\end{table}

In Table \ref{tab:imageperf} the output per image is shown; its true label, and the number and fraction of cells in this image that is correctly assigned to the true class. For most images the performance is  good, but for some images the procedure completely fails: in particular on images 2, 9, 10, 17 and 23. These images are from the ``coarse speckled'' and ``fine speckled'' classes, so these results are in line with the confusion matrices. 

We also investigated the effect of other cell label combining rules on the image level performance. The mean and product rule showed a slight improvement in performance, which is consistent with previous results that these combiners are more robust than majority voting~\cite{kittler1998combining}. However, the improvement was not very significant, suggesting that other measures are necessaary to further increase the performance.

\subsection{Proposed: Image level}

Here we show experiments for the multiple instance learning approach, where each image is represented by the set of quantile levels (10\%, 11\%, 50\% and 100\%). A logistic classifier is trained and tested on the images as a whole. Therefore, in the test phase, we first obtain image labels, and then propagate these labels to the corresponding cells.

In this task, we also evaluated several classifiers (1-norm SVM, SVM, logistic and 
nearest neighbor) and the logistic classifier performed the best on image level.

\begin{table}[!ht]
\caption{The confusion matrices and classification performances using the logistic classifier on image level on the six cell classes: \cen:``centromere'', \coa:``coarse speckled'', \cyt=``cytoplasmatic'', \fin:``fine speckled'', \hom:``homogeneous'', \nuc:``nucleolar''. Top row shows results on the predefined train-test split, the bottom row the results on 28-fold cross validation.}
\label{tab:imbased}
\small
\begin{tabular}{cc}
\hline
\multicolumn{2}{c}{Split in predefined train and test set}\\
\hline
cell level evaluation & image level evaluation \\
\begin{tabular}{l|*{6}{c}}
& \multicolumn{6}{c}{estimated class} \\

true & \cen & \coa & \cyt & \fin & \hom & \nuc \\ 
 \hline 
 \cen & 84 & 0 & 0 & 0 & 0 & 65 \\
      \coa & 0 & 33 & 0 & 68 & 0 & 0 \\
      \cyt & 0 & 0 & 51 & 0 & 0 & 0 \\
      \fin & 0 & 0 & 0 & 51 & 63 & 0 \\
      \hom & 0 & 0 & 0 & 61 & 119 & 0 \\
      \nuc & 0 & 0 & 0 & 0 & 0 & 139 \\

\end{tabular}
&

\begin{tabular}{l|*{6}{c}}
& \multicolumn{6}{c}{estimated class} \\
     true & \cen & \coa & \cyt & \fin & \hom & \nuc \\ 
 \hline

      \cen & 2 & 0 & 0 & 0 & 0 & 1 \\
      \coa & 0 & 1 & 0 & 2 & 0 & 0 \\
      \cyt & 0 & 0 & 2 & 0 & 0 & 0 \\
      \fin & 0 & 0 & 0 & 1 & 1 & 0 \\
      \hom & 0 & 0 & 0 & 1 & 1 & 0 \\
      \nuc & 0 & 0 & 0 & 0 & 0 & 2 \\

\end{tabular} \\

 65.0\% correct &   64.3\% correct \\
\hline
\multicolumn{2}{c}{28-fold cross-validation}\\
\hline
\begin{tabular}{l|*{6}{c}}
& \multicolumn{6}{c}{estimated class} \\
     true & \cen & \coa & \cyt & \fin & \hom & \nuc \\ 
 \hline 
      \cen & 357 & 0 & 0 & 0 & 0 & 0 \\
      \coa & 0 & 161 & 0 & 49 & 0 & 0 \\
      \cyt & 0 & 0 & 109 & 0 & 0 & 0 \\
      \fin & 0 & 48 & 0 & 97 & 63 & 0 \\
      \hom & 0 & 0 & 0 & 42 & 288 & 0 \\
      \nuc & 66 & 0 & 0 & 0 & 0 & 175 \\
\end{tabular}
&
\begin{tabular}{l|*{6}{c}}
& \multicolumn{6}{c}{estimated class} \\
     true & \cen & \coa & \cyt & \fin & \hom & \nuc \\ 
 \hline 
      \cen & 6 & 0 & 0 & 0 & 0 & 0 \\
      \coa & 0 & 4 & 0 & 1 & 0 & 0 \\
      \cyt & 0 & 0 & 4 & 0 & 0 & 0 \\
      \fin & 0 & 1 & 0 & 2 & 1 & 0 \\
      \hom & 0 & 0 & 0 & 1 & 4 & 0 \\
      \nuc & 1 & 0 & 0 & 0 & 0 & 3 \\
\end{tabular} \\

81.6\% correct &  82.1\% correct \\

\end{tabular}
\end{table}

The top row of Table \ref{tab:imbased} shows the results on the predefined train test split. Although the cell level performance improves slightly, the image level performance worsens. In other words, images with more cells are classified correctly more often, but in total less images are classified correctly. With just 14 training slides, the classifier is not able to generalize well to the predefined test set. 

The second row of Table \ref{tab:imbased} shows the results of the cross-validation. Here the performances are significantly improved as opposed to the train test split. The classifier is able to form a better model of the six classes because more training slides are available. The improvement in cell level performance is especially large, over 20\%. This demonstrates that cells that were previously misclassified, a classified correctly due to the other cells present in the image. 

We believe that the cross-validation results are a better indicator of the performance than the results on the predefined train test split. In our investigation, we have experimented with alternative features and classifiers, and in all cases, high accuracies on the train test split corresponded with an overtrained classifier, and low performances in cross-validation. 

Table \ref{tab:imageperf2} shows the performances of our approach on each individual image. Because the classification is done on bag level and the instance labels are propagated, it is only possible to classify 0\% or 100\% of the cells per image correctly. The images  that are misclassified (2, 4, 12, 15 and 18) are somewhat different from the images in Table \ref{tab:imageperf} that also had very poor performance: 2, 9, 10, 17 and 23. This suggests that combining the two approaches might lead to further improvement in accuracy.

\begin{table}[ht]
\centering
\caption{Classification performance for each individual image. From left to right: the image number, the true class label, the number of cells that is correctly classified, and the fraction of cells that is correctly classified.}
\label{tab:imageperf2}
\small
\begin{tabular}{rlrr}
 & true label & corr. & fraction\\
\hline
 1 & \homtxt &  61 &  100.0\% \\ 
  2 & \fintxt &   0 &  0.0\% \\ 
  3 & \centxt &  89 &  100.0\% \\ 
  4 & \nuctxt &   0 &  0.0\% \\ 
  5 & \homtxt &  47 &  100.0\% \\ 
  6 & \coatxt &  68 &  100.0\% \\ 
  7 & \centxt &  56 &  100.0\% \\ 
  8 & \nuctxt &  56 &  100.0\% \\ 
  9 & \fintxt &  46 &  100.0\% \\ 
 10 & \coatxt &  33 &  100.0\% \\ 
 11 & \coatxt &  41 &  100.0\% \\ 
 12 & \coatxt &   0 &  0.0\% \\ 
 13 & \centxt &  46 &  100.0\% \\ 
 14 & \centxt &  63 &  100.0\% \\ 
\end{tabular}
\begin{tabular}{rlrr}
 & true label & corr. & fraction\\
\hline
 15 & \fintxt &   0 &  0.0\% \\ 
 16 & \centxt &  38 &  100.0\% \\ 
 17 & \coatxt &  19 &  100.0\% \\ 
 18 & \homtxt &   0 &  0.0\% \\ 
 19 & \centxt &  65 &  100.0\% \\ 
 20 & \nuctxt &  46 &  100.0\% \\ 
 21 & \homtxt &  61 &  100.0\% \\ 
 22 & \homtxt & 119 &  100.0\% \\ 
 23 & \fintxt &  51 &  100.0\% \\ 
 24 & \nuctxt &  73 &  100.0\% \\ 
 25 & \cyttxt &  24 &  100.0\% \\ 
 26 & \cyttxt &  34 &  100.0\% \\ 
 27 & \cyttxt &  38 &  100.0\% \\ 
 28 & \cyttxt &  13 &  100.0\% \\ 
\end{tabular}
\end{table}

Even with the improved results, there is confusion between ``fine speckled'' and ``coarse speckled''. We examined the images that are misclassified. The two ``fine speckled'' images 2 and 15 are of very low quality and the cells in them are nearly indistinguishable. The ``coarse'' speckled image 12, looks similar to ``fine speckled'' images in the dataset, at least to an untrained eye. Perhaps the extracted Gabor features do not sufficiently capture the differences between the two classes and features designed with the help of an expert could help to reduce the class overlap.

\section{Conclusions} \label{sec:concl}

In this paper we propose a quantile representation to represent a set of feature vectors. This representation characterizes a collection of feature vectors, or a bag of instances in the terminology of multiple instance learning.
For each bag, several quantile values for each of the features are computed. This results in a vector with fixed length for bags with a variable number of instances, a representation that can be used in any standard classifier. Although this representation ignores the correlation between the features, it characterizes the marginal feature distributions of the bags well.

This quantile representation is applicable for the classification of an image containing several HEp-2 cells. All of the cells in an image share the same label, and therefore all cells are informative for the image label. Experiments show reasonably good classification performance on image level, which also results in a better performance on cell level. Unfortunately, there is still confusion between the classes ``fine speckled'' and ``coarse speckled'', and between ``fine speckled'' and ``homogeneous''. For non-experts like the authors, these classes are indeed very difficult to distinguish.
 
%The advantage of using all cells from one image in classification is, that for the training of the classifier only image-level labels have to be provided. This alleviates the need for an expert to label individual cells (or instances in a bag).

A downside of the current quantile representation is that the quantile levels have to be chosen appropriately. When no prior knowledge is available on what quantile level might be informative, a quantile level selection is required. It would be interesting to investigate whether this can be done by visually inspecting the distributions of the classes for each feature.

%On the other hand, by inspecting the distributions of the classes for each feature, one may notice if bags from some classes have long tails, or lack a particular mode in a distribution. This may be used to infer what the informative quantiles might be. In a sense, this is similar to feature selection, but instead of selecting whole features, only the more informative parts of each feature distribution are used. 

The proposed representation is a general and intuitive representation that can be readily applied to other classification problems, and might therefore be helpful for other automatic diagnostic systems.

%TODO: more conclusions?

\bibliographystyle{plain}    % 
\bibliography{publications}

\end{document}